*

# A Beam-Segmenting Polar Format Algorithm Based on Double PCS for Video SAR Persistent Imaging

Jiawei Jiang, Yinwei Li, Shaowen Luo, Ping Li and Yiming Zhu

*Abstract*—Video synthetic aperture radar (SAR) is attracting more attention in recent years due to its abilities of high resolution, high frame rate and advantages in continuous observation. Generally, the polar format algorithm (PFA) is an efficient algorithm for spotlight mode video SAR. However, in the process of PFA, the wavefront curvature error (WCE) limits the imaging scene size and the 2-D interpolation affects the efficiency. To solve the aforementioned problems, a beam-segmenting PFA based on principle of chirp scaling (PCS), called BS-PCS-PFA, is proposed for video SAR imaging, which has the capability of persistent imaging for different carrier frequencies video SAR. Firstly, an improved PCS applicable to video SAR PFA is proposed to replace the 2-D interpolation and the coarse image in the ground output coordinate system (GOCS) is obtained. As for the distortion or defocus existing in the coarse image, a novel sub-block imaging method based on beam-segmenting fast filtering is proposed to segment the image into multiple sub-beam data, whose distortion and defocus can be ignored when the equivalent size of sub-block is smaller than the distortion negligible region. Through processing the sub-beam data and mosaicking the refocused sub-images, the full image in GOCS without distortion and defocus is obtained. Moreover, a three-step MoCo method is applied to the algorithm for the adaptability to the actual irregular trajectories. The proposed method can significantly expand the effective scene size of PFA, and the better operational efficiency makes it more suitable for video SAR imaging. The feasibility of the algorithm is verified by the experimental data.

## I. INTRODUCTION

AS a novel observation technology, video synthetic aperture radar (SAR) overcomes the low frame rate shortcomings of traditional SAR [1]-[3]. Video SAR forms a sequence of SAR images by persistent surveillance of the region of interest (ROI) and processing of the received echoes. When the time to form each image is short enough, the imaging process is similar to playing video [4]. These features make video SAR an excellent choice for continuous tracking of ground targets. For a certain azimuth resolution, the frame rate of video SAR is proportional to the carrier frequency [5]. To achieve video-like observation of ground targets such as vehicles, video SAR requires a resolution of 0.2m and the frame rate needs to exceed 5Hz [6]. Therefore, video SAR usually operate in the terahertz (THz) band and works in circular spotlight mode [7]–[9].

Two commonly used algorithms for circular spotlight mode are back projection algorithm (BPA) and polar format algorithm (PFA). BPA can perform high-precision imaging under arbitrary flight trajectory, but pixel-by-pixel processing of the echo signal makes its calculation burden too large [10]. The modified version, such as the fast backprojection algorithm (FBP) [11] and the fast factorized backprojection algorithm (FFBP) [12], has improved the efficiency of BPA. However, this comes at the cost of losing imaging accuracy and is still less efficient than frequency-domain algorithms.

The approximation of planar wavefront (APW) leads to higher computational efficiency of PFA [12][13], making it more suitable for video SAR image formation. However, there are still several problems when applying PFA directly to video SAR. First, the 2-D interpolation in polar format transform (PFT) affects the computational efficiency of PFA [15]. In addition, the APW simplifies the process of the algorithm but also introduces the wavefront curvature error (WCE). The constant and linear terms error of WCE lead to geometric distortion of the image, the quadratic terms error leads to image defocus, and the cubic and higher-order terms error affects the main-lobe and side-lobe region, respectively [16], [17]. Thus, WCE affects the imaging quality of the PFA, which is not negligible for the desired high-resolution video SAR images. Meanwhile, the directions of every frame image are different due to the change of azimuth angle [18], while video SAR images in the ground output coordinate (GOCS) are desired to facilitate the target location and tracking.

In [19], a parameter-adjusting autoregistration PFA is proposed for linear spotlight mode video SAR. The method replaces the range interpolation in PFT by changing the system

*This research was supported in part by the Natural Science Foundation of Shanghai (Grant No.21ZR1444300), in part by the National Natural Science Foundation of China (Grant No. 61988102, 61731020), in part by the National key R&D Project of China (Grant No. 2018YFF01013003). (Corresponding author: Yinwei Li)

J. Jiang and S. Luo are with the School of Optoelectronic Information and Computer Engineering, University of Shanghai for Science and Technology, Shanghai200093,China(202310315@st.usst.edu.cn,213330652@st.usst.edu.cn).

Y. Li, P. Li and Y. Zhu are also with Terahertz Technology Innovation Research Institute, Terahertz Spectrum and Imaging Technology Cooperative Innovation Center, University of Shanghai for Science and Technology, Shanghai 200093, China (e-mail: liyw@usst.edu.cn, liping@usst.edu.cn, ymzhu@usst.edu.cn).

parameters in real time, but the SAR system structure is more expensive and difficult to implement compared to the fixed-parameter system. In [20], the 2-D interpolation in PFT is replaced by range and azimuth chirp-Z transform (CZT), and an image-domain 2-D interpolation is used to complete the process of geometric distortion correction and coordinate system alignment. But the ignorance of quadratic phase error (QPE) leads to image defocus and the existing image interpolation still affects the efficiency. In [21], a quadtree beam-segmenting PFA is proposed to solve the image defocus for the wide-swath SAR. The method moves the wide-beam to each sub-block center and obtain sub-beam data by using filtering method, and the full image without defocus is formed through mosaicking the sub-images. However, the method requires a full-size complex multiplication and imaging process for each sub-beam, which leads to a significant increase in computational load as the number of sub-beams increases. In [22], the RZPFA rotates whole wide-beam to multiple refocusing points and processed them to obtain full images, then intercepts and mosaics the defocus-negligible sub-images to obtain the final full image. However, multiple imaging processes for the refocused whole wide-beam make it computationally expensive, so it is only applicable for ROI but not the whole large scene of video SAR.

Considering the problems of applying PFA to video SAR, a beam-segmenting PFA based on the principle of chirp scaling (PCS), named BS-PCS-PFA, is proposed for persistent imaging of the circular spotlight mode video SAR. Firstly, to avoid the 2-D interpolation in PFT, based on the PCS of linear frequency modulated (LFM) signal [23], the range PCS (RPCS) and azimuth PCS (APCS) applicable to video SAR PFA are derived in detail. Compared to CZT, PCS also involves only FFT and complex multiplication, but has a shorter processing chain and a more efficient FFT [24], [25]. Then, a beam-segmenting fast filtering method is proposed to decompose the wide beam data into a set of narrow sub-beam data, the center point of each sub-block center is treated as the reference center for motion compensation (MoCo), and each sub-beam data only contains information of corresponding sub-block. When the equivalent sub-block size is smaller than the distortion-negligible region (DiR), the effect of distortion and defocus within the sub-block can be ignored. Through processing each sub-beam data and mosaicking the refocused sub-image to real position, the full SAR image without distortion and defocus under GOCS is obtained. Moreover, considering the effect of motion errors, a three-step MoCo method is applied to adapt the irregular shape of trajectory. Since the imaging processes of the sub-blocks are independent of each other, it is suitable for parallel processing in hardware such as FPGAs to further accelerate the imaging efficiency. The main contributions of this paper are:

(1) An improved PCS applicable to video SAR PFA is proposed to replace the wave number domain interpolation and improve the efficiency;
(2) A novel sub-block imaging method based on digital spotlight is proposed for the efficient resolution of wavefront curvature error compensation in video SAR PFA;

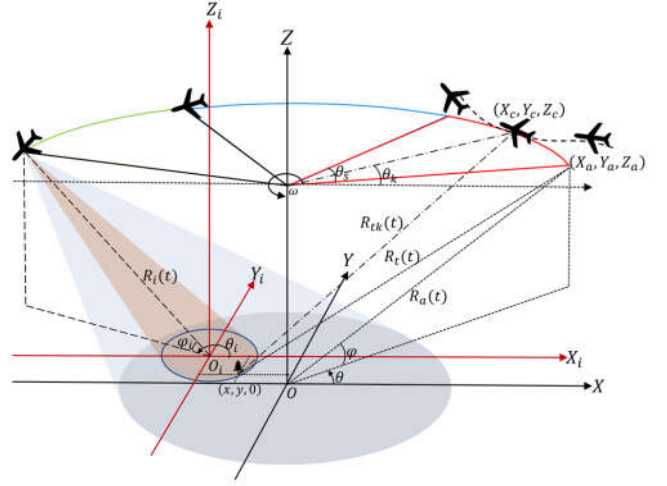

**Fig. 1.** Imaging geometry of video SAR.

(3) The proposed method can efficiently improve the efficiency and expand the effective scene size of PFA.

The article is organized as follows. In Section II, the imaging geometry and signal model of video SAR are described in detail. In Section III, the impact of WCE for video SAR PFA is analyzed. In Section IV, a beam-segmenting PFA based on double PCS for video SAR persistent imaging is proposed. In Section V, the algorithm is validated by experimental analysis. Finally, the conclusions are given in Section VI.

## II. SIGNAL MODEL OF VIDEO SAR

### A. Geometry Definition

The imaging geometry model of video SAR is shown in Fig. 1. The center of the ground scene coincides with the coordinate center $O$ of the Cartesian coordinate system $X$-$Y$-$Z$. The beam illuminates the ROI during the radar flight for persistent observation of the ground target. The velocity is $v$, the flight radius is $R_s$. $\theta_k$ represents the central azimuth of the $k$th frame sub-aperture and $\theta_s$ is the rotated angle of each frame. $R_a(t)$ denotes the slant range between radar platform position $(X_a, Y_a, Z_a)$ and the scene center $O$, $\theta$ denotes the azimuth angle and $\varphi$ denotes elevation angle of the radar platform, respectively. And the slant range $R_t(t)$ between platform and ground point $p(x,y)$ is expressed as:

$$R_t(t) = \sqrt{(X_a - x)^2 + (Y_a - y)^2 + Z_a^2} \qquad (1)$$

Let $t = 0$ denotes the aperture center moment of slow time $t$, and $(X_c, Y_c, Z_c) = (X_a, Y_a, Z_a)|_{t=0}$ denotes the aperture center position of the platform, then the instantaneous slant range is:

$$R_{tk}(t) = \sqrt{(X_c - x)^2 + (Y_c - y)^2 + Z_c^2} \qquad (2)$$

When the system parameters and azimuth resolution $\rho_a$ are constant, the imaging frame rate $F_r$ of video SAR is related to the system carrier frequency and sub-aperture overlap ratio $\omega$:

$$F_r = \frac{2\rho_a v}{(1-\omega)R_a c} f_c \qquad (3)$$

Assuming that the average slant range of video SAR is $1000m$ and the velocity is $50m/s$. To achieve the imaging frame rate of 5Hz, the carrier frequency of video SAR needs to exceed 207GHz. Thus, video SAR usually works in higher frequency to satisfy the requirements of frame rate when $\omega$ is zero. Considering the effect of atmospheric attenuation on the signal [26], the frequency of 220 GHz is a reasonable choice due to the relatively small attenuation. Meanwhile, to work in low frequency bands, there exists one method to increase the image frame rate by increasing the overlap ratio $\omega$ such as [27]. Although the synthetic aperture time does not change in this mode, i.e., the delay between the video SAR image and the actual scene, it still has some use [28], [29]. Therefore, the proposed method in this paper will be applicable to different carrier frequencies.

*B. Signal Model*

The LFM is transmitted, then the echo signal of arbitrary ground target $p(x, y, 0)$ can be described as:

$$s_R(\tau,t) = \sigma \cdot \text{rect}\left[\frac{\tau - \frac{2R_t(t)}{c}}{T_r}\right]$$
$$\cdot \exp\left\{j2\pi\left[f_c\left(\tau - \frac{2R_t(t)}{c}\right) + \frac{1}{2}K_r\left(\tau - \frac{2R_t(t)}{c}\right)^2\right]\right\} \quad (4)$$

where $\sigma$ is the scattering intensity, $\tau$ is fast time, $T_r$ is pulse width, $f_c$ is carrier frequency and $K_r$ is the chirp rate of LFM.

Assuming that the actual position of platform is accurately measured by the difference global positioning system and inertial measurement unit (DGPS&IMU). Then, the 1st MoCo is applied by multiplying the scene center echo as reference. After that the dechirped signal prepared for polar format storage can be described as:

$$s_{if}(\tau,t) = \sigma \cdot \text{rect}\left[\frac{\tau - \frac{2R_t}{c}}{T_r}\right]$$
$$\cdot \exp\left\{j\frac{4\pi K_r}{c}\left(\tau - \frac{2R_a}{c}\right)\Delta R - j\frac{4\pi f_c}{c}\Delta R + j\frac{4\pi K_r}{c^2}\Delta R^2\right\} \quad (5)$$

where $\Delta R$ is the differential distance between the scene center point $O$ and the target $p$:

$$\Delta R = \sqrt{(X_a - x)^2 + (Y_a - y)^2 + Z_a^2} - R_a(t) \quad (6)$$

The range fast Fourier transform (FFT) is performed on the dechirped signal:

$$S(f_\tau,t) = \sigma \cdot T_r \text{sinc}\left[T_r\left(f_\tau + \frac{2K_r \Delta R}{c}\right)\right]$$
$$\cdot \exp\left\{-j\frac{4\pi f_c}{c}\Delta R - j\frac{4\pi f_c}{c^2}\Delta R^2 - j\frac{4\pi f_\tau}{c}\Delta R\right\} \quad (7)$$

where $f_\tau$ is fast time frequency, $\text{sinc}[\cdot]$ is sinc function.

The dechirped signal includes residual video phase (RVP) term and envelope term, which correspond to the second and third term in the phase of (7), respectively. These two terms lead to side flaps suppression and image defocus. An effective compensation method is to multiply a compensation function in range frequency domain [30]:

$$S_c(f_\tau) = \exp\left(-j\frac{\pi f_\tau^2}{K_r}\right) \quad (8)$$

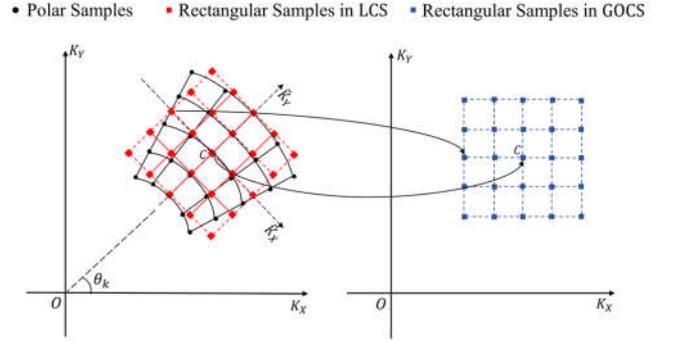

Fig. 2. Polar format transformation of wavenumber signal.

Multiplying (7) with (8) and then implement range IFFT:

$$s_{ic}(\tau,t) = \sigma \cdot \text{rect}\left[\frac{\tau - \frac{2R_a(t)}{c}}{T_r}\right]$$
$$\cdot \exp\left\{-j\left[\frac{4\pi f_c}{c} + \frac{4\pi K_r}{c}\left(\tau - \frac{2R_a(t)}{c}\right)\right]\Delta R\right\} \quad (9)$$

Suppose that $K_R = \frac{4\pi f_c}{c} + \frac{4\pi K_r}{c}\left(\tau - \frac{2R_a(t)}{c}\right)$ denotes the spatial wavenumber, $K_X(K_R,t) = K_R \cos\varphi(t)\cos\theta(t)$ and $K_Y(K_R,t) = K_R \cos\varphi(t)\sin\theta(t)$ are range and azimuth wavenumber, respectively, where the azimuth angle $\theta = arctan\left(\frac{Y_a}{X_a}\right)$ and the elevation angle $\varphi = arcsin\left(\frac{Z_a}{R_a}\right)$. Then, based on the APW, the differential distance $\Delta R$ can be simplified as:

$$\Delta R_p(t) \approx -x\cos\varphi(t)\cos\theta(t) - y\cos\varphi(t)\sin\theta(t) \quad (10)$$

To express simplicity, amplitude information is ignored because it hardly affects the imaging algorithm, then the signal in wavenumber-time domain has the form of:

$$S(K_R,t) = \exp\{j[xK_X(K_R,t) + yK_Y(K_R,t)]\} \quad (11)$$

$S(K_R,t)$ is uniformly distributed in $K_R - t$ domain, yet is non-uniformly distributed in $K_X - K_Y$ due to the change of azimuth angle $\theta(t)$. As shown by the black samples in Fig. 2, the distribution of $S(K_R,t)$ in wavenumber domain takes the form of a polar format. In order to take advantage of the efficiency of 2D-FFT, a 2-D interpolation should be implemented in PFA to obtain the rectangular uniform samples. Then the 2D resampled wavenumber $(\widetilde{K}_{X_r}, \widetilde{K}_{Y_r})$ after the line-of-sight polar interpolation (LOSPI) [18] can be expressed as:

$$\widetilde{K}_{X_r} = K_R \cos\varphi \cos\theta_k - K_{X_c}\tan(\theta - \theta_k)$$
$$\widetilde{K}_{Y_r} = K_R \cos\varphi \sin\theta_k + K_{Y_c}\tan(\theta - \theta_k) \quad (12)$$

where $(\widetilde{K}_{X_r}, \widetilde{K}_{Y_r})$ corresponds to the red rectangular samples in Fig. 2, and $K_{X_c}$ and $K_{Y_c}$ represent the range and azimuth wavenumber of point C:

$$K_{X_c} = \frac{4\pi}{c} f_c \cos\varphi \cos\theta_k$$
$$K_{Y_c} = \frac{4\pi}{c} f_c \cos\varphi \sin\theta_k \quad (13)$$

In this case, however, the change in azimuth angle will result in a rotation of the final image. Therefore, as shown in Fig. 2, a rotation operation should be implemented to convert $(\widetilde{K}_{X_r}, \widetilde{K}_{Y_r})$ to the blue rectangular sample $(K_{X_r}, K_{Y_r})$ in GOCS:

$$\begin{bmatrix} K_{X_r} - K_{X_c} \\ K_{Y_r} - K_{Y_c} \end{bmatrix} = \begin{bmatrix} \cos\theta_k & \sin\theta_k \\ -\sin\theta_k & \cos\theta_k \end{bmatrix} \begin{bmatrix} \tilde{K}_{X_r} - K_{X_c} \\ \tilde{K}_{Y_r} - K_{Y_c} \end{bmatrix} \quad (14)$$

Based on (12)-(14), the expressions of $(K_{X_r}, K_{Y_r})$ can be obtained as:

$$K_{X_r} = \frac{4\pi}{c}(f_\tau + f_c \cos\theta_k)\cos\varphi$$
$$K_{Y_r} = \frac{4\pi}{c}f_c[(\theta - \theta_k)\sec^2\theta_k + \sin\theta_k]\cos\varphi \quad (15)$$

Finally, a 2D-FFT is implemented to convert the rectangular samples into a focused image. And the focused image in GOCS is obtained, where the position coordinates of targets do not change with azimuth angle.

### III. WAVEFRONT CURVATURE ERROR ANALYSIS OF VIDEO SAR PFA

The signal distribution of PFA has a relatively simple form due to the APW, which simplifies the imaging process of algorithm. The approximation can be considered to be approximately valid in the case of long range or small scenes. However, APW also results in geometric distortion and image defocus when the approximation is not met, such as short range or large scenes. The geometric distortion performs as the mapping from the actual position $(x, y)$ to the distortion position $(x^*, y^*)$. Image defocus is caused by the quadratic terms phase error. The WCE is proportional to the distance between target and beam center. Therefore, it is necessary to compensate the WCE for high-resolution video SAR system.

Performing Taylor series expansion of the differential range at the aperture center moment $t = 0$, the constant term of $\Delta R$ in (7) and $\Delta R_p$ in (10) are:

$$\Delta R^{(0)} = \Delta R|_{t=0} = R_{tk} - R_a$$
$$\Delta R_p^{(0)} = \Delta R_p|_{t=0} = \frac{x^* X_c + y^* Y_c}{-R_a} \quad (16)$$

and the linear term of $\Delta R$ and $\Delta R_p$ are:

$$\Delta R^{(1)} = \left.\frac{\partial \Delta R}{\partial t}\right|_{t=0} = \frac{(xY_c - yX_c)\theta_s}{2R_{tk}}$$
$$\Delta R_p^{(1)} = \left.\frac{\partial \Delta R_p}{\partial t}\right|_{t=0} = \frac{(x^* Y_c - y^* X_c)\theta_s}{2R_a} \quad (17)$$

Let $\Delta R^{(0)} = \Delta R_p^{(0)}$ and $\Delta R^{(1)} = \Delta R_p^{(1)}$, yielding:

$$x^* X_c + y^* Y_c = R_a^2 - R_a R_{tk}$$
$$x^* Y_c - y^* X_c = \frac{R_a(xY_c - yX_c)}{R_{tk}} \quad (18)$$

Then, the distortion mapping of ground target can be represented as:

$$x^*(x, y; k) = \frac{(R_a - R_{tk})\cos\theta_k}{\cos\varphi} - \frac{(xX_c - xY_c)\sin\theta_k}{R_{tk}\cos\varphi}$$
$$y^*(x, y; k) = \frac{(R_a - R_{tk})\sin\theta_k}{\cos\varphi} + \frac{(yX_c - xY_c)\cos\theta_k}{R_{tk}\cos\varphi} \quad (19)$$

Obviously, the distortion mapping is related to both the position of ground target and the frame of video SAR. Let $r_d$ represents the offset distance of the target, then the offset in the $k$th frame can be expressed as:

$$r_d^k(x, y) = \sqrt{(x - x^*)^2 + (y - y^*)^2} \quad (20)$$

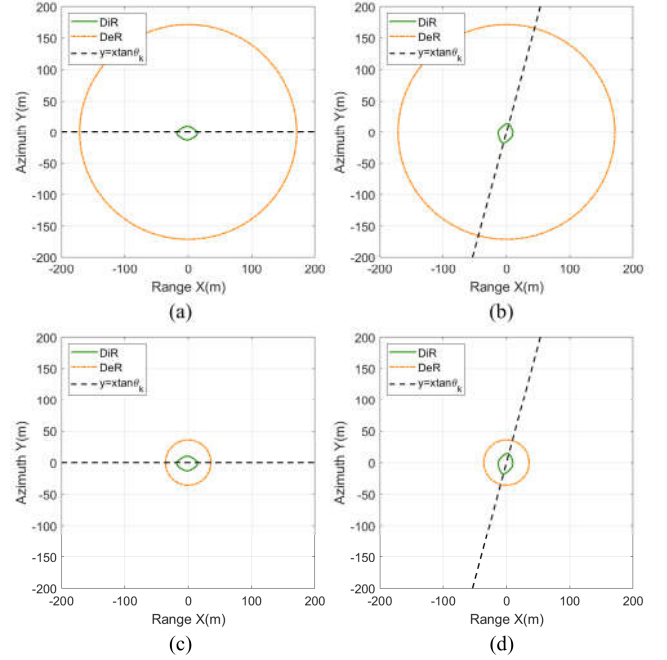

**Fig. 3.** DiR and DeR of video SAR PFA for different carrier frequency and azimuth angle. (a) $f_c = 220\text{GHz}$, $\theta_k = 0°$; (b) $f_c = 220\text{GHz}$, $\theta_k = 75°$; (c) $f_c = 9.6\text{GHz}$, $\theta_k = 0°$; (d) $f_c = 9.6\text{GHz}$, $\theta_k = 75°$.

By limiting the offset $r_d^k(x, y)$ to the matching resolution $\rho_x$, then the DiR is restricted as:

$$D_i^k(x, y) = \{(x, y) | r_d^k(x, y) \leq \rho_x\} \quad (21)$$

The quadratic and cross Taylor series terms can be computed using the same method, and the detail results are given by [31]. The quadratic term leads to the image defocus, which is proportional to the distance from the point target to the scene center. As a result, the QPE limits the scene size and the phase error corrections are required once the scene size exceeds the limitations. However, the phase error of $\pi/2$ or $\pi/4$ in time-to-frequency conversion is usually considered to be negligible, and [20] gives the maximum scene radius $r_{max} = \rho_a\sqrt{2R_a/\lambda}$. Then, the defocus-negligible region (DeR) is given by:

$$D_e^k = \{|R_{scene}| \leq r_{max}\} \quad (22)$$

When the average distance $\overline{R_a} = 500m$ and the resolution $\rho_x = 0.2m$, the DiR and DeR of video SAR for different carrier frequency and azimuth angle are showed in Fig. 3. Fig. 3(a) and (b) give the DiR and DeR of the carrier frequency $f_c = 220\text{GHz}$ with $\theta_k = 0°$ and $\theta_k = 75°$. Fig. 3(c) and (d) give the DiR and DeR of the carrier frequency $f_c = 9.6\text{GHz}$ with $\theta_k = 0°$ and $\theta_k = 75°$. It can be seen that the DiR is only related to the target position and not related to the carrier frequency. And the size and shape of DiR is constant and symmetric about $y = \tan\theta_k$. The DeR of 220 GHz and 9.6 GHz are 171.32m and 35.78m respectively, which means that the THz video SAR has a larger DeR. Therefore, the corrections for both distortion and defocus are necessary for low-frequency video SAR, while the effect of WCE on THz video SAR consists only geometric distortion as the imaging scene is rather small.

## IV. BEAM-SEGMENTING PFA BASED ON DOUBLE PCS

The high-frame rate feature of video SAR requires a short imaging time, otherwise a time delay that accumulates with the observation will be generated. As described in Section III, only the reference point and its surrounding area can ignore the effects of geometric distortion and image defocus. In order to avoid the above-mentioned issues, a beam-segmenting PFA based on PCS and sub-block imaging strategy is proposed to realize more efficient imaging of video SAR.

*A. Polar Format Algorithm Based on Double PCS*

The property of LFM makes it possible to implement the resampling without interpolation, which is called PCS. The initial version of time scaling and frequency scaling [32] is shown in Fig. 4(a) and (b), respectively. For example, the signal $p(t)$ is transformed into $p(\xi t + \beta)$ after PCS, where $\xi$ and $\beta$ represent the scaling factors. In this section, to void the wavenumber domain 2-D interpolation, the PCS-PFA, which is based on range PCS (RPCS) and azimuth PCS (APCS) are derived in detail.

As described in the previous sections, the 2D resampling operations can be expressed as converting polar samples $(K_X, K_Y)$ to the resampled rectangular samples $(K_{X_r}, K_{Y_r})$ in GOCS, which is well demonstrated in the transformation from black samples to blue samples in Fig. 2. Generalizing $f_\tau$ to $\xi_r f_\tau + \beta_r$, then the range wavenumber is expressed as:

$$K'_X = \frac{4\pi}{c}(f_c + \xi_r f_\tau + \beta_r)\cos\varphi \cos\theta \quad (23)$$

Let $K'_X = K_{X_r}$, based on (15) and (23), the range scaling factors is derived as:

$$\xi_r = \frac{1}{\cos\theta}$$
$$\beta_r = \frac{(\cos\theta_k - \cos\theta)}{\cos\theta}f_c \quad (24)$$

The RPCS is derived from the frequency-domain scaling, which is shown by the gray region in Fig. 4(a). Define at the start of processing:

$$P(f) = \mathbf{F}_\tau[s_{ic}(\tau, t)] \quad (25)$$

where $\mathbf{F}_\tau$ represents range FFT.

Next, define the $H_1(f_\tau)$ as follows:

$$H_1(f_\tau) = \exp\left(\frac{-j\pi^2 f_\tau^2}{\eta_r}\right) = \exp\left(\frac{j\pi f_\tau^2}{K_r}\right) \Rightarrow \eta_r = -\pi K_r \quad (26)$$

Then the processing up to point **a** is expressed as:

$$s_{if}(\tau, t) = \mathbf{F}_\tau^{-1}\{\mathbf{F}_\tau[s_{ic}(\tau, t)] \cdot H_1(f_\tau)\} \quad (27)$$

It can be noted that $\mathbf{F}_\tau[s_{ic}(\tau, t)]$ is equivalent to $P(f_\tau)$, and the original dechirped signal $s_{if}(\tau, t)$ without RVP elimination is obtained at point **a**. This helps avoid a range IFFT operations for the dechirp-on-receive system, which reduces the processing chain and improves computational efficiency.

Proceeding with the processing up to point **b**, the range scaled signal is:

$$s_d\left\{\xi_r\left[\tau - \frac{2R_a}{c} - \frac{f_c(\xi_r - 1)}{K_r}\right] + \frac{2R_a}{c} + \frac{f_c(\xi_r - 1)}{K_r}, t\right\} \quad (28)$$

Unlike the traditional interpolation-based PFA, here the removal of RVP and range compression are not performed additionally, but are embedded in the RPCS process. The output range scaled signal $S_R(f_\tau, t)$ at point **c** can be obtained as:

$$S_R(f_\tau, t) = \exp\left\{-j\frac{4\pi \cos\varphi}{c}(f_\tau + f_c \cos\theta_k)(x + y\tan\theta)\right\} \quad (29)$$

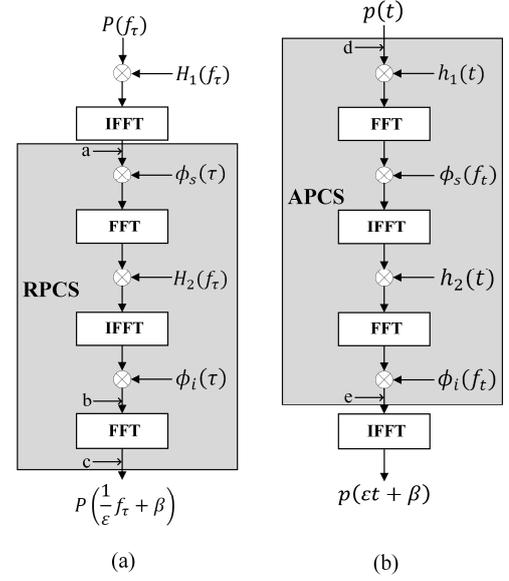

**Fig. 4.** The principle of chirp scaling. (a) Frequency scaling and RPCS (b) Time scaling and APCS.

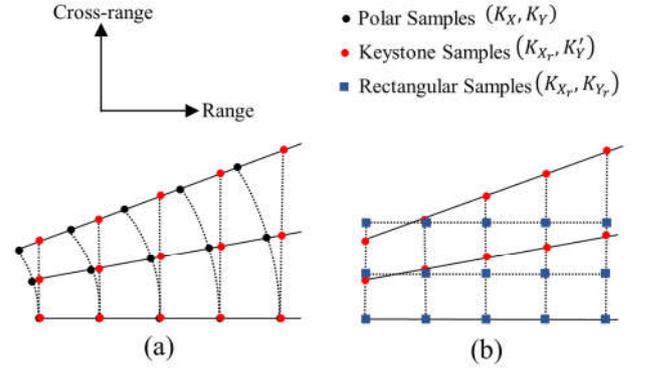

**Fig. 5.** Resampling from polar grid to rectangular grid. (a) Range resampling. (b)Azimuth resampling.

The system parameters of RPCS are:

$$\phi_s(\tau) = \exp\left\{-j\eta_r(1 - \xi_r)\left(\tau - \frac{2R_a}{c}\right)^2\right\}$$
$$H_2(f_\tau) = \exp\left\{-j\frac{\pi}{\xi_r K_r}f_\tau^2\right\} \cdot \exp\left\{j2\pi K_r f_c \frac{\xi_r - 1}{\xi_r}\right\}$$
$$\phi_i(\tau) = \exp\{j\eta_r(\xi_r - \xi_r^2)\}$$
$$\cdot \exp\left\{\left[\tau + \frac{(\xi_r - 1)f_c}{K_r \xi_r} - \frac{2R_a}{c}\right] - j2\pi\frac{\beta_r}{\xi_r}\tau\right\} \quad (30)$$

Fig. 5 shows a more detailed schematic of the sample transformation, the wavenumber-domain samples after RPCS are transformed from polar grid to keystone grid, and the azimuth wavenumber is:

$$K_Y' = \frac{4\pi}{c}\cos\varphi\,(f_\tau + f_c\cos\theta_k)$$
$$\cdot [\tan\theta_k + (\theta - \theta_k)\sec^2\theta_k] \quad (31)$$

Next is the keystone-to-rectangular transformation, which is performed by APCS. Generalizing $\theta$ to $\xi_a\theta + \beta_a$, then it can be expressed as:

$$K_Y'' = \frac{4\pi}{c}\cos\varphi\,(f_\tau + f_c\cos\theta_k)$$
$$\cdot [\tan\theta_k + (\xi_a\theta + \beta_a - \theta_k)\sec^2\theta_k] \quad (32)$$

Let $K_Y'' = K_{Y_r}$, based on (15) and (32), the azimuth scaling factors is derived as:

$$\xi_a = \frac{f_c \cos^2\theta_k}{f_\tau + f_c\cos\theta_k}$$
$$\beta_a = \frac{f_\tau\left(\theta_k - \frac{1}{2}\sin 2\theta_k\right) + \theta_k(\cos\theta_k + \cos^2\theta_k)}{f_\tau + f_c\cos\theta_k} \quad (33)$$

The process of APCS is shown in the gray region in Fig. 4(b), since the output of RPCS has been dechirped in azimuth, a rechirped-operation should be implemented at point **d** in Fig. 4(b), which can be implemented by defining $h_1(t)$ as follows:

$$h_1(t) = \exp(j\eta_a t^2) = \exp(j\pi K_a t^2) \Rightarrow \eta_a = \pi K_a \quad (34)$$

where $K_a = \frac{2v^2 \sin^2\varphi}{\lambda R_a}$ is the Doppler rate at the aperture center.

As shown in Fig. 5(b), APCS is equivalent to an azimuth scaling operation, which is equivalent to converting the keystone samples into the rectangular samples. The output azimuth scaled signal at point **e** can be obtained as follow:

$$S_{RA}(f_\tau, f_t) = \mathbf{F}_t[S_R(f_\tau, \xi_a t)] \quad (35)$$

where $\mathbf{F}_t$ represents azimuth Fourier transform.

The system parameters of APCS are:

$$h_1(t) = \exp(j\eta_a t^2)$$
$$\phi_s(f_t) = \exp\left(j\pi\frac{\xi_a - 1}{\xi_a K_a}f_t^2\right)$$
$$h_2(t) = \exp(-j\xi_a\eta_a t^2)$$
$$\phi_i(f_t) = \exp\left(j\pi\frac{\xi_a - 1}{\xi_a^2 K_a}f_t^2 + j2\pi\frac{\beta_a}{\xi_a}f_t\right) \quad (36)$$

The 2D-FFT in PFA is completed by the range FFT in the last term of RPCS and the azimuth FFT in the last term of APCS, respectively. Therefore, the whole process of PCS-PFA ends at point **e** in Fig. 4(b) and the focused SAR image in GOCS is obtained. Different from the traditional frequency-domain chirp scaling algorithm (CSA) [33], which is only work in straight-line mode, the PCS-PFA is also suitable for persistent curve spotlight mode of video SAR.

*B. Beam-segmenting Fast Filtering Based on Digital Spotlight*

For the distortion or defocus in coarse image of PCS-PFA, the proposed beam-segmenting fast filtering method segments the image into multiple sub-beam data based on the principle of digital spotlight [34]. Each sub-beam data only contains information of corresponding sub-block. When the equivalent sub-block size is smaller than the DiR, the effect of distortion and defocus within the sub-block can be ignored. As shown in Fig. 1, the process is actually equivalent to moving the beam to illuminate a small area near the scene center $O_i(X_i, Y_i)$. Establishing a new coordinate system $X_iY_iZ_i$ with $O_i$ as the reference point, then the slant range from the APC to the scene center in the new coordinate system is:

$$R_i(t) = \sqrt{(X_a - X_i)^2 + (Y_a - Y_i)^2 + Z_a^2} \quad (37)$$

where the elevation angle $\varphi_i = \arcsin\left(\frac{Z_a}{R_i(t)}\right)$ and the azimuth angle $\theta_i = \arctan\left(\frac{Y_a - Y_i}{X_a - X_i}\right)$.

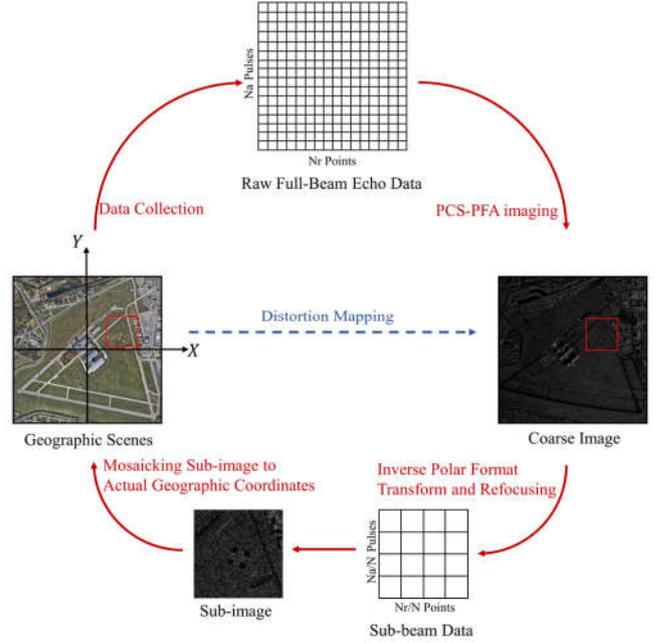

**Fig. 6.** Process of sub-image.

The process of sub-image is shown in Fig. 6. First, the ground scene is divided into sub-blocks in the geographic coordinate system. Next, the distortion position $(X_i^*, Y_i^*)$ of sub-block center can be calculated through the distortion mapping. Then intercept the sub-images with sub-block center and ensure that the size is within the DiR. From Fig. 3, it is known that DiR is always smaller than the DeR, so the effect of WCE can be ignored inside the sub-block. The segmentation criterion in this paper is equal division, i.e., selecting equal subregions shaped as a rectangle, uniformly from the upper left to the lower right in the full image. Assuming that the full scene size is $R_{size} \times R_{size}$, the segmentation level is $N$ and the sub-image size is $W_r \times W_r$. To make the targets inside the sub-block all lie within the DiR, $W_r$ should satisfy $W_r \leq \frac{\sqrt{2}}{2}\gamma$, where $\gamma$ is a fixed value and can be denoted as $\gamma = \mathbf{r}_d(|x - x^*| = \rho_x, 0)$. Then the criterion of the segmentation level $N$ should be referred to:

$$N = \left\lfloor \frac{R_{size}}{W_r} \right\rfloor \quad (38)$$

where $\lfloor\cdot\rfloor$ represents rounding down operation. Considering the possible distortion or defocus in coarse image, the selected sub-image should be slightly larger than the sub-block size to avoid energy leakage. Besides a small overlap exists between adjacent sub-images to ensure the continuity of full image.

The polar resampling in PFA is an endomorphism mapping, which has phase-preserving property [35]. Therefore, the corresponding sub-beam echo data in the original phase history domain can be obtained by implementing an IPFT to the wavenumber domain sub-beam data, which can be decomposed into inverse azimuth resample and inverse-range resample. The range and azimuth sampling numbers reduce to $1/N$ after sub-image processing, i.e., the number of sampling points of the four domains of signals $\tau, f_\tau, t, f_t$ are reduced to $1/N$, which are denoted by $\tau_d$, $f_{\tau d}$, $t_d$, $f_{td}$, respectively. These ensure complexity of algorithm not increasing too much. Using the down-sampled parameters, the PCS can be used to replaced 2-D interpolation in the inverse polar format transform (IPFT), which can be called inverse azimuth PCS (IAPCS) and inverse range PCS (IRPCS). IAPCS and IRPCS are the inverse processes of RPCS and APCS, respectively. Moreover, the system functions need to be redefined and the system functions of IRPCS are:

$$\phi_i(f_{td}) = \exp\left(j\pi \frac{\xi_a - 1}{\xi_a^2 K_a} f_{td}^2 + j2\pi \frac{\beta_a}{\xi_a} f_{td}\right)$$
$$h_2(t_d) = \exp(-j\xi_a \eta_a t_d^2)$$
$$\phi_s(f_{td}) = \exp\left(j\pi \frac{\xi_a - 1}{\xi_a K_a} f_{td}^2\right)$$
$$h_1(t_d) = \exp(j\eta_a t_d^2) \tag{39}$$

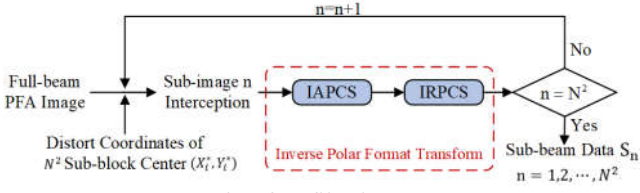

**Fig. 7.** Beam-segmenting fast filtering.

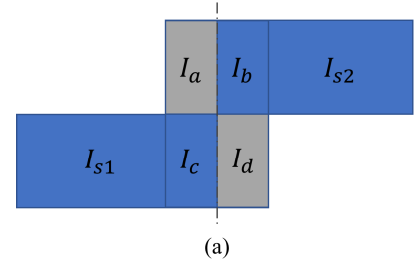

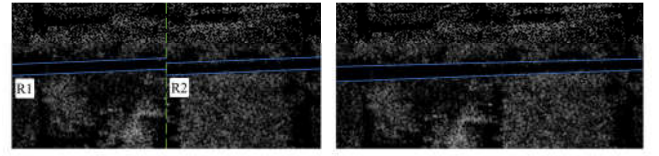

**Fig. 8.** Strategy of image mosaicking.

The system functions of IAPCS are:

$$\phi_i(\tau_d) = \exp\left\{j\eta_r(\xi_r - \xi_r^2)\left[\tau_d + \frac{(\xi_r - 1)f_c}{K_r \xi_r} - \frac{2R_a}{c}\right]\right\}$$
$$H_2(f_{\tau d}) = \exp\left\{-j\frac{\pi}{\xi_r K_r} f_{\tau d}^2\right\} \cdot \exp\left\{j2\pi K_r f_c \frac{\xi_r - 1}{\xi_r}\right\}$$
$$\phi_s(\tau_d) = \exp\left\{-j\eta_r(1 - \xi_r)\left(\tau_d - \frac{2R_a}{c}\right)^2\right\} \tag{40}$$

Then, the downsampled sub-beam data after IRPCS and IAPCS can be expressed as:

$$S_{dn} = \sum_{n=1}^{N^2} \sigma_n \cdot \exp\left\{j\frac{4\pi}{c}(f_c + f_{\tau d})[R_a(t) - R_t(t)]\right\} \tag{41}$$

The sub-beam data $S_{dn}$ is recovered to the original phase history domain. Then, the 2nd MoCo is implemented to each sub-beam data with reference to its corresponding sub-block center, i.e., multiplied by the following reference function:

$$S_{ref.n} = \sum_{n=1}^{N^2} \exp\left\{j\frac{4\pi}{c}(f_c + f_{\tau d})(R_i(t) - R_a(t))\right\} \tag{42}$$

After the refocusing and 2nd MoCo, the 2-D dechirped signal containing only sub-block information can then be obtained:

$$S_n = \sum_{n=1}^{N^2} \sigma_n \cdot \exp\left\{j\frac{4\pi}{c}(f_c + f_{\tau d})[R_i(t) - R_t(t)]\right\} \tag{43}$$

The process discarding redundant information and focusing only on the ROI information can be called beam-segmenting fast filtering. The schematic diagram of beam-segmenting fast filtering is shown in Fig. 7.

Considering the limited accuracy of devices, the signal processed by DGPS&IMU-based MoCo method may still exists residual motion error, therefore, the Map-Drift (MD) [36] algorithm based 3rd MoCo is applied to compensate the residual motion error. Moreover, the multi-squint (MS) [37] processing approach and weighted phase curvature autofocus (WPCA) [38] can be applied to appliable to compensate the residual motion error of in different modes, such as InSAR.

*C. Sub-block Imaging Based on PFA and Image Mosaicking*

After the three-step MoCo, the motion error of sub-block center is accurately compensated. Since the size of sub-block is small enough, the APW is satisfied, and the sub-beam data can be processed by PFA according to the new sub-block center, i.e., implementing the range resampling, azimuth resampling and 2-D FFT to obtain the refocused sub-image. Likewise, the resampling interpolation can also be replaced based on PCS. However, the wavenumber-domain 2-D resampling process is not always necessary, in some case it is possible to process signal distributed in polar format without affecting the image quality. In [39], the guidelines are given:

$$|X_\varepsilon| = \frac{2\rho_r \rho_a \cos\varphi}{\lambda}$$
$$|Y_\varepsilon| = \frac{2\rho_a^2 \cos\varphi}{\lambda} \tag{44}$$

In fact, in the case of resolution matching, the following relationship always holds:

$$\gamma < |X_\varepsilon| = |Y_\varepsilon| \tag{45}$$

Therefore, under the condition that the sub-block size $W_r$ always meet (44), the signal $S_n(f_{\tau d}, t)$ can be processed directly while ignoring the 2-D resampling of the sub-beam data, which improves the efficiency of the algorithm.

After the process of all sub-blocks, the full image is obtained through mosaicking all sub-images according to the actual position. However, a displacement may exist between two adjacent sub-images when directly mosaicking them.

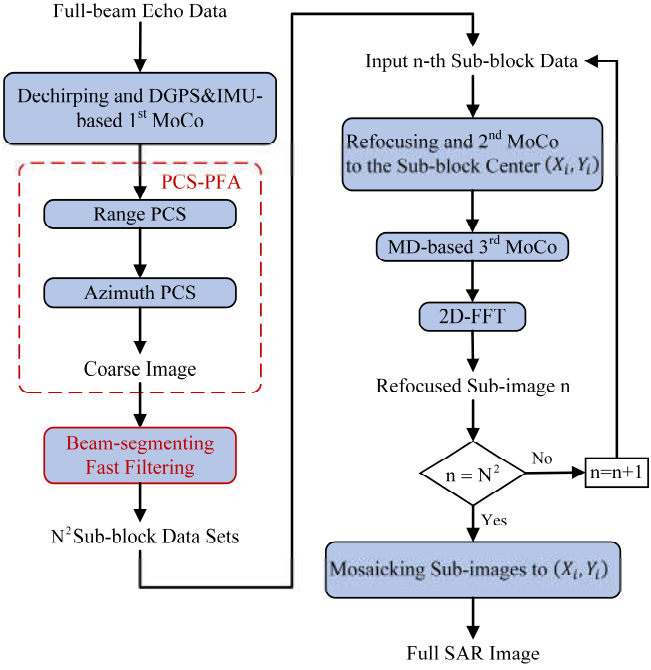

**Fig. 9.** Flow chart of the proposed BS-PCS-PFA.

Therefore, a certain overlap is retained when sub-images are divided, the adjacent sub-images can be easily mosaicked by overlapping region correlation registration and radiation correction [40], [41], which ensure the spectrum of point targets at the edge of sub-image is uninterrupted.

Let $I_{s1}(x,y)$ and $I_{s2}(x,y)$ denote the two adjacent sub-images described in Fig. 8(a), and the displacement is $(\Delta x, \Delta y)$, then the relationship between them can be expressed as:
$$I_{s2}(x,y) = I_{s1}(x + \Delta x, y + \Delta y) \quad (46)$$
By using the shift property of the FFT, the relationship between $I_{s1}(x,y)$ and $I_{s2}(x,y)$ in the Fourier domain can be expressed as:
$$\frac{I_{s2}(K_x,K_y)}{I_{s1}(K_x,K_y)} = \exp\{j[K_x\Delta x + K_y\Delta y]\} \quad (47)$$
Then the displacement $(\Delta x, \Delta y)$ can be obtained by calculating the maximum of the IFFT of (47):
$$(\Delta x, \Delta y) = \underset{(x,y)}{\mathrm{argmax}}\{\mathrm{IFFT}[\exp\{j[K_x\Delta x + K_y\Delta y]\}]\}$$
$$= \underset{(x,y)}{\mathrm{argmax}}\{\delta(x + \Delta x, y + \Delta y)\} \quad (48)$$
Fig. 8(b) and Fig. 8(c) show the adjacent sub-images before and after the correlation registration, respectively, which indicates that the problem of dislocation is solved.

After that the full SAR image without distortion, defocus and dislocation can be obtained, and the complete flowchart of the proposed BS-PCS-PFA is shown in Fig. 9.

*D. Computation Load Analysis*

In this section, the BPA, PFA-LOSPI, UCSA [20] and our method are discussed for the comparison of computation load. A complex multiplication requires six floating point operations (flops). A 1-D FFT of length $N_i$ requires $5N_i log_2 N_i$ flops and a 2-D FFT of $N_i \times N_i$ requires $10N_i^2 log_2 N_i$ flops.

The BPA performs interpolation to every pulse, followed by coherent accumulation along slow time, without any approximation. The computational load of the BPA is:
$$C_1 = 5N_aN_r log_2(N_r) + 6N_aN_r$$
$$+ 8N_aN_xN_y + 5MN_a log_2(MN_r) \quad (46)$$
where $N_x$ and $N_y$ are the grids number and $M$ is the length of the interpolation kernel.

For PFA-LOSPI, the main processing includes range FFT, 2-D interpolation for PFT, 2-D IFFT and image interpolation for distortion correction. Therefore, the computational load of the PFA-LOSPI is:
$$C_2 = 10N_aN_r log_2(N_r) + [4M^2 + 10M]N_aN_r \quad (47)$$
where $N_a$ and $N_r$ are the number of azimuth and range points, respectively.

The UCSA uses range CZT and azimuth CZT to replace 2D-IFFT and 2-D interpolation of PFT. The computation of CZT is equivalent to three times $2N$ point length FFT, so the computational load of the UCSA is:
$$C_3 = 10N_aN_r log_2(N_r) + 30N_aN_r log_2 2(N_r + N_a)$$
$$+ 4(2M - 1)N_aN_r \quad (48)$$
For the PCS-PFA, the main processing includes the range PCS and azimuth PCS, while the RVP compensation and Fourier imaging process are concluded. The computational load of the PCS-PFA is:
$$C_4 = 30N_aN_r log_2 N_a + 42N_aN_r \quad (49)$$
For the proposed beam-segmenting PFA based on double PCS, the main processing includes RPCS and APCS in the PCS-PFA, followed by IAPCS and IRPCS in the beam-segmenting fast filtering, and a finally 2D-FFT. Since the beam-segmenting fast filtering processes sub-beams of smaller size, the computational complexity does not increase much. The computational load of the BS-PCS-PFA is:
$$C_5 = 15N_aN_r log_2\left(\frac{N_a^2 + N_r^2}{N}\right)$$
$$+ 5N_aN_r log_2\left(\frac{N_aN_r}{N^2}\right) + 90N_aN_r \quad (50)$$
Assuming that $N_x = N_y = 1024$, $N_r = N_a = 1024$, the length of interpolation kernel $M = 16$ and the segmentation level $N = 8$. Then the ratio of the computational load of the BS-PCS-PFA and the other methods are:
$$\begin{cases} p_1 = \dfrac{C_5}{C_1} = 0.045 \\ p_2 = \dfrac{C_5}{C_2} = 0.322 \\ p_3 = \dfrac{C_5}{C_3} = 0.736 \\ p_4 = \dfrac{C_5}{C_4} = 1.257 \end{cases} \quad (51)$$

The reason for the ratio $p_3$ between our method and UCSA stems from two main points. 1) Compared to the mandatory requirement of zero augmenting in the CZT, PCS holds a shorter processing chain and more efficient FFT. 2) The combination for distortion correction and coordinate system alignment makes the proposed method more efficient. This indicates that the proposed method has higher computational efficiency compared to the previous methods. Furthermore, the computational load PCS-PFA is lower than BS-PCS-PFA, so PCS-PFA is a better choice for scenarios that both meet long slant range, small scene size and high frequency. However, the image quality of PCS-PFA will be affected when any condition is not met, the harsh condition makes BS-PCS-PFA an efficient and more versatile choice.

## V. Experiment Analysis

### A. Point Target Simulation Results

To verify the effectiveness of the proposed method, the point target and extended target simulation are conducted in spotlight mode video SAR. The radar system parameters are listed in Table I. A short average slant range $\bar{R}_a = 500m$ is used to highlight the effect of geometric distortion. The scene size is $130m \times 130m$, point targets are distributed in a rectangular range of $100m \times 100m$, and the interval between adjacent target points is 10m. Therefore, from (21), DiR is 23.8m. From (22), it can be calculated that the DeR for $f_c = 220GHz$ and $f_c = 9.6GHz$ are 171.3m and 35.7m respectively. Fig. 10 shows the imaging results while carrier frequency is 220GHz and azimuth angle $\theta_k = 0°$. It is known that all targets are within the DeR and the effect of QPE can be ignored. The magenta circle in Fig. 10(a) shows the real position of the point targets and the red circle shows the theoretical distortion position through calculated with (19). As shown in Fig. 10(b), the distortion is present in the imaging result of PCS-PFA, which coincides with the theoretical analysis. And the solid green line represents the DiR at $\theta_k = 0°$, where the distortion is controlled within unit resolution. As shown in Fig. 10(b), the distortion is present in the imaging result of PCS-PFA, which coincides with the theoretical analysis. And the solid green line represents the DiR at $\theta_k = 0°$, where the distortion is controlled within unit resolution. Next, the sub-images are intercepted according to the distortion position of the sub-blocks. Through using IPFT and second PFA, the sub-image is refocused and mosaiced into the real position in the pre-defined empty matrix, as shown in Fig. 10(d). After processing each sub-image, the full distortion-free image of BS-PCS-PFA can be obtained as shown in Fig. 10(c). It is obvious that all the targets are located at the real position. The imaging result of traditional PFA-LOSPI [42] is also shown in Fig. 10(e). It is clearly seen that the distortion is present and similar to PCS-PFA. The imaging result of standard BPA is shown in Fig. 10(f) for comparison.

To verify the performance of the proposed method at different azimuth angles, the method is used to focus the image under the following parameters: $f_c = 220GHz$, $\theta_k = 75°$. Fig.11(a) shows the distortion mapping when $\theta_k = 75°$. The PCS-PFA image shown in Fig. 11(b) still lies in the GOCS without rotation, verifying the accuracy of the PCS-PFA using double PCS. It can be seen that the distortion position coincides with the distortion mapping at $\theta_k = 75°$ and the green solid line in Fig. 11(b) shows the DiR with symmetry about $y = x \tan 75°$. The BS-PCS-PFA image is shown in Fig. 11(c) and all the targets are located at the real position. Fig. 11(e) and

TABLE I
System Parameters

| Parameters | Values |
| --- | --- |
| Carrier frequency | 9.6/220GHz |
| Bandwidth | 1.2GHz |
| Reference slant range | 500m |
| Grazing Angle | π/4 |
| Velocity | 30m/s |
| Resolution | 0.125m |

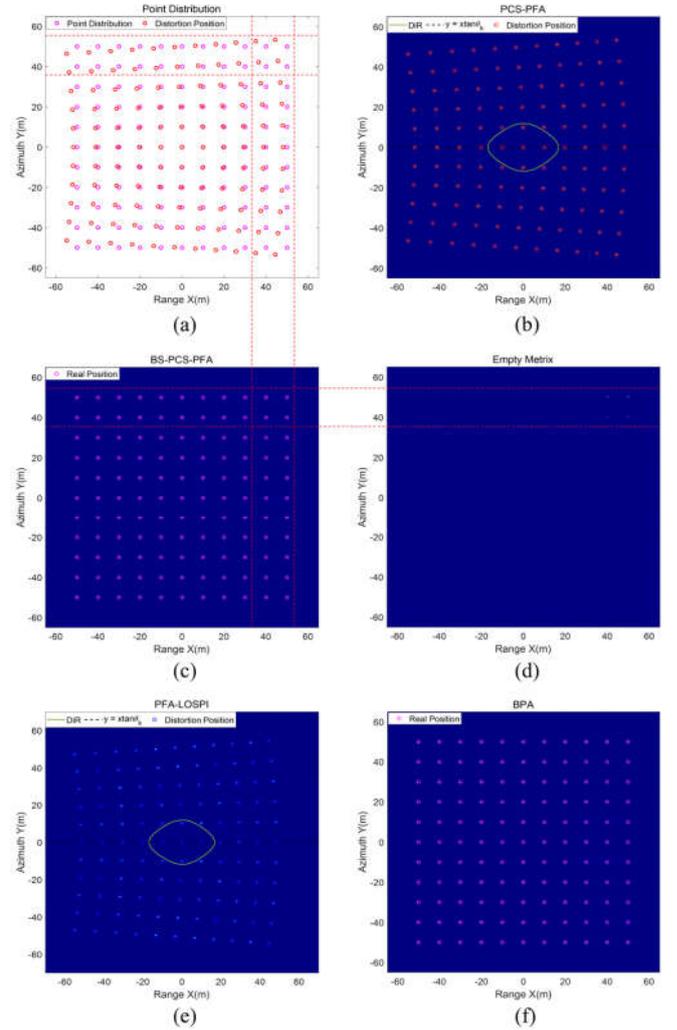

**Fig. 10**. Imaging results of in 220GHz while $\theta_k = 0°$. (a)Points distribution and distortion position. (b)PCS-PFA.(c)BS-PCS-PFA. (d)Empty matrix. (e)PFA-LOSPI. (f)BPA.

Fig.11(f) shows the imaging results of PFA-LOSPI and BPA. The distortion relationship between PFA-LOSPI and (19) is a rotation of $\theta_k$, which can be expressed as:
$$\tilde{x}^*(x,y;k) = x^* \cos\theta_k + y^* \sin\theta_k$$
$$\tilde{y}^*(x,y;k) = -x^* \cos\theta_k + y^* \cos\theta_k \quad (52)$$

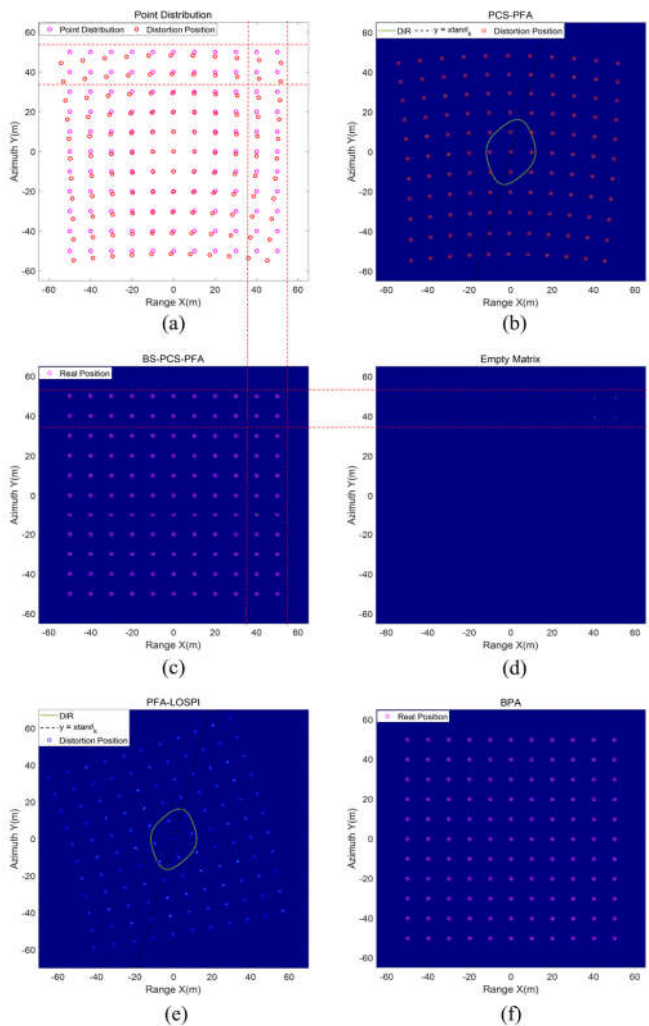

**Fig. 11.** Imaging results in 220GHz while $\theta_k = 75°$. (a)Points distribution. (b)PCS-PFA. (c)BS-PCS-PFA. (d)Empty matrix. (e)PFA-LOSPI. (f)BPA.

which is equal only at $\theta_k = 0°$, so it is clearly not acceptable for the requirements of target localization.

To quantitatively analyze the imaging performance of point targets, the range and azimuth profiles of the point $P(50,50)$ obtained by different algorithms at an azimuth angle of 0° and 75° are shown in Fig.12, and the measured impulse response width (IRW), peak side lobe ratio (PSLR) and integral side lobe ratio (ISLR) are listed in Table II. It can be seen that the image quality of PCS-PFA and BS-PCS-PFA is similar to PFA-LOSPI and BPA. So the proposed method can be considered to meet the system requirements of video SAR. The measured positions of point P(50,50) with three methods are given in Table III. For PFA-LOSPI and PCS-PFA, there is a distortion of several meters when $\theta_k = 0°$. Furthermore, the distortion of PFA-LOSPI increases to over a dozen meters when $\theta_k = 75°$. In contrast, the measured position coordinates of BS-PCS-PFA and BPA at different angles are considered to be accurate enough, which facilitates the target localization of video SAR.

To verify the performance of the method for different carrier frequency, the following system parameters are used: $f_c = 9.6GHz$ when $\theta_k = 0°$ and $\theta_k = 75°$. Compared with the THz band, a data multiplexing rate of 90.4% is required to meet the 5Hz imaging frame rate under such conditions, which affects the real-time performance of the system, while THz video SAR basically does not require data multiplexing. Fig. 13 (a) and (d) shows the imaging results with the PFA-LOSPI method. The existing distortion with respect to (19) is a rotation of $\theta_k$, and the image scene is slightly enlarged because some points are out of range due to the image rotation. The zoom-in imaging results of three points indicate that the point within the DeR is well focused, while the points outside the DeR are defocus. Fig. 13 (b) and (e) shows the imaging results with the PCS-PFA method. It can be seen that the distortion position and DiR coincide with the corresponding results in THz band, but the DeR is relatively smaller. Compared with the PFA-LOSPI, the images are both under the GOCS whatever the azimuth angle is. Besides, the main flap of the point spread function (PSF) is rotating when $\theta_k = 75°$. Fig. 13 (c) and (f) shows the results of BS-PCS-PFA. Compared with the previous two methods, the points targets are located in the real position without distortion, and the corresponding zoom-in imaging results show that all point targets are well focused. These demonstrate the persistent imaging ability of the proposed method for video SAR at different carrier frequencies.

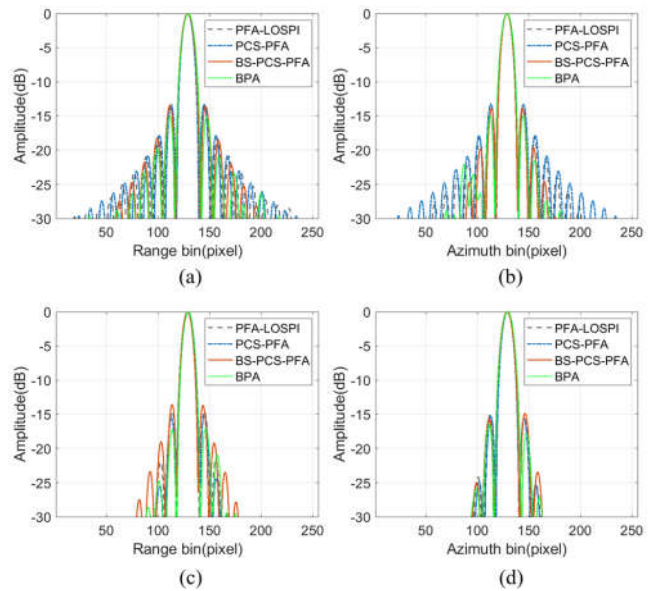

Fig. 12. Image profiles of point target P. (a) Range profiles with $\theta_k = 0°$. (b) Azimuth profiles with $\theta_k = 0°$. (c) Range profiles with $\theta_k = 75°$. (d) Azimuth profiles with $\theta_k = 75°$.

TABLE II
Imaging Quality Inspection of Target P

| $\theta_k = 0°$ | Resolution(m) | | PSLR(dB) | | ISLR(dB) | |
|---|---|---|---|---|---|---|
| | Range | Azimuth | Range | Azimuth | Range | Azimuth |
| PFA-LOSPI | 0.128 | 0.128 | -13.30 | -13.20 | -26.11 | -27.17 |
| PCS-PFA | 0.126 | 0.128 | -13.27 | -13.22 | -29.71 | -29.15 |
| BS-PCS-PFA | 0.129 | 0.126 | -13.41 | -13.80 | -30.98 | -28.85 |
| BPA | 0.129 | 0.129 | -14.85 | -14.96 | -30.35 | -28.20 |
| $\theta_k = 75°$ | Resolution(m) | | PSLR(dB) | | ISLR(dB) | |
| | Range | Azimuth | Range | Azimuth | Range | Azimuth |
| PFA-LOSPI | 0.130 | 0.131 | -14.46 | -15.13 | -29.71 | -30.22 |
| PCS-PFA | 0.131 | 0.129 | -15.04 | -15.31 | -35.41 | -35.63 |
| BS-PCS-PFA | 0.129 | 0.130 | -14.55 | -14.84 | -34.15 | -35.76 |
| BPA | 0.131 | 0.132 | -17.07 | -16.41 | -32.97 | -33.50 |

TABLE III
Geometry Position of Point Target P

| | $\theta_k = 0°$ | $\theta_k = 75°$ |
|---|---|---|
| PFA-LOSPI | (44.47,53.36)m | (56.69,-38.54)m |
| PCS-PFA | (44.28,53.30)m | (51.52,44.47)m |
| BS-PCS-PFA | (50.02,49.99)m | (50.00,49.98)m |
| BPA | (50.01,50.02)m | (50.02,49.98)m |

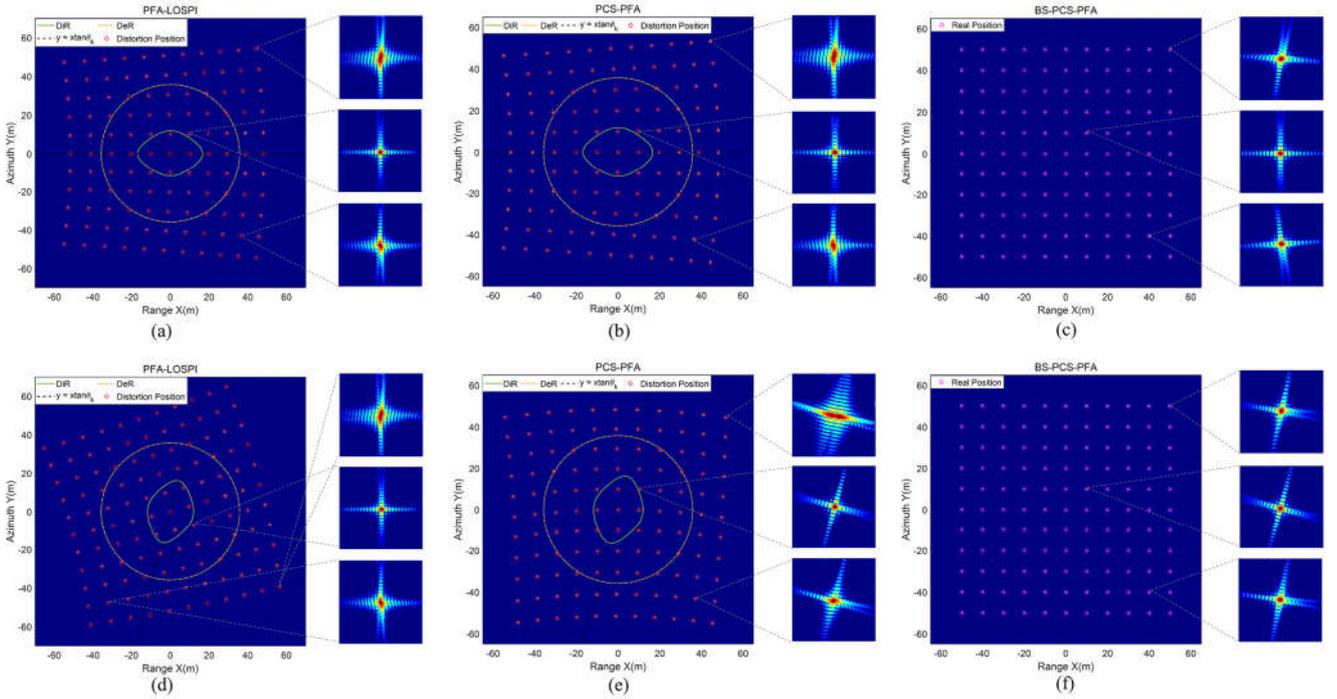

**Fig.13.** Imaging Results of ground points in 9.6GHz. (a)-(c) Imaging results of PFA-LOSPI, PCS-PFA and BS-PCS-PFA while $\theta_k = 0°$. (d)-(f) Imaging results of PFA-LOSPI, PCS-PFA and BS-PCS-PFA while $\theta_k = 75°$.

TABLE IV
Average Elapsed Time Comparison

| PFA-LOSPI (without DC) | PFA-LOSPI (with DC) | PCS-PFA | BS-PCS-PFA | BPA |
|---|---|---|---|---|
| 0.56s | 1.35s | 0.48s | 0.58s | 45.02s |

The average imaging time of above algorithms is compared to verify the computational efficiency. The simulation environment adopts MATLAB 2021a in Windows 11 (64-bit operating system and 16G memory size) with AMD Ryzen 5 5600H CPU, and Cubic interpolation is used for distortion correction (DC) of PFA-LOSPI. The average single frame elapsed times of above methods are listed in Table IV, it shows the great advantage of the frequency-domain PFA over the time-domain BPA, and the runtime advantage of the proposed method over the conventional PFA and BPA is verified. Moreover, considering the aperture overlap strategy, in order to achieve a 5Hz imaging frame rate, BPA requires an overlap ratio of 99.5% while BS-PCS-PFA is 65.5%. However, the imaging frame rate is not exactly the same as the single-frame image formation time, and the BPA based on aperture overlap still responds to the echo data before a certain time and does not reflect the observed scene information in real time.

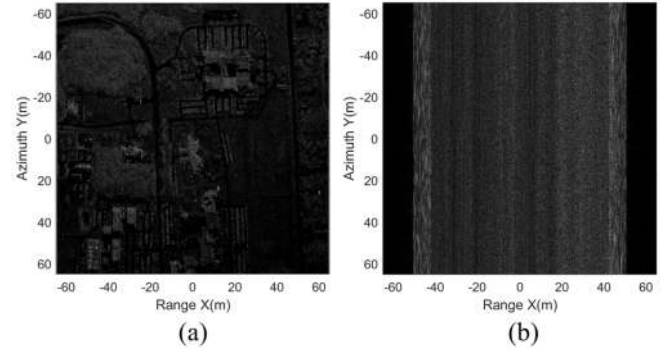

Fig.14. (a) Original SAR image of the extended target simulation. (b) Range compressed image.

*B. Extended Target Simulation*

To further validate the performance of the proposed method, an extended target simulation is performed. The input image is shown in Fig. 14(a), which is obtained from the website of Air

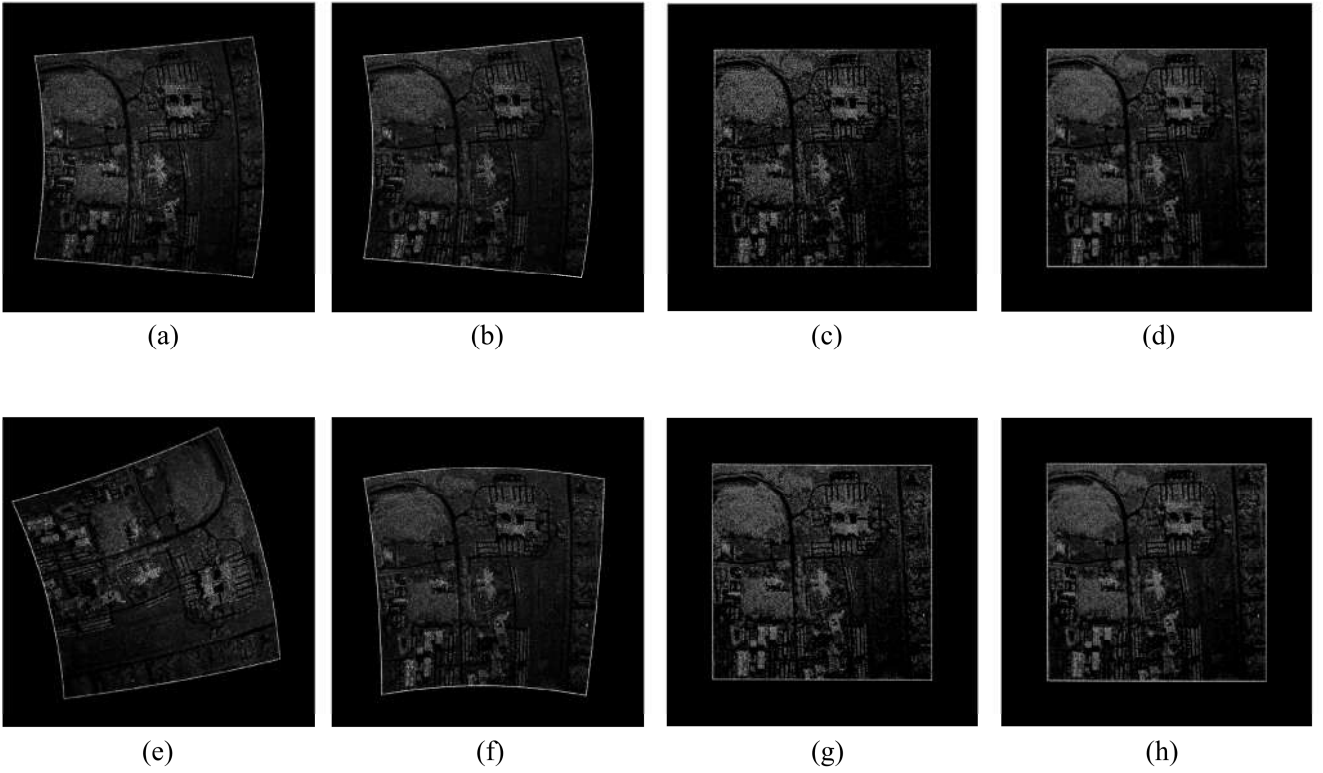

**Fig.15.** Imaging results of extended target. (a)-(d) Imaging results processed by PFA-LOSPI, PCS-PFA, BS-PCS-PFA and BPA when $\theta_k = 0°$. (e)-(h) Imaging results processed by PFA-LOSPI, PCS-PFA, BS-PCS-PFA and BPA when $\theta_k = 75°$.

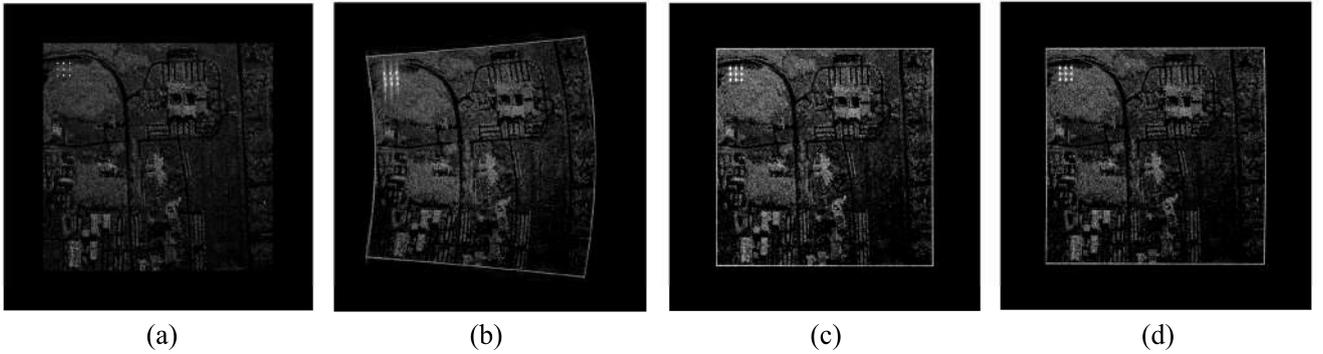

**Fig.16.** Imaging results of strong scattering points. (a) Original input image with strong scattering points. (b)-(d) Imaging result of PFA-LOSPI, BS-PCA-PFA and BPA.

Force Research Laboratory (AFRL) [43]. The echo data is obtained by time domain simulation method [44], and the pre-processed range compressed image is shown in Fig.14 (b). The same scene size and system parameters were used as in the previous point target simulation, which means that the DiR and DeR of the image scene remain unchanged.

Fig. 15 shows the imaging results of above methods at a carrier frequency of 220 GHz and different azimuth angles of 0° and 75°, where all scenes have been resized to the same size. It can be seen that all the methods can obtain the well-focused SAR images, verifying the ability of the PFA operating in the THz band to focus well over a wide size. However, due to the LOSPI of the method, the PFA-LOSPI image rotates with azimuthal angle $\theta_k$ and the distortion introduced by the linear phase error distorts the imaging results into a sector-like image. For the PCS-PFA, the imaging results are always under the GOCS, and only distortion is present in the image. According to (38) and the sub-image processing method, the whole PCS-PFA image is divided into $8 \times 8 = 64$ sub-images, with an adjacent overlap of 8.6m, each sub-image is processed by the beam segmenting fast filtering and refocused by PFA again. The imaging results of BS-PCS-PFA show that the distortion is corrected and the image is under the GOCS, where no additional processing of coordinate system alignment is required. The imaging results of BPA are also given in the last column for comparison.

In order to more intuitively illustrate the compensation effect of the proposed method on WCE, a strong scattering point target set in the edge region with a frequency of 9.6 GHz is tested. Fig. 16(a) shows the original input image. The scattering points are in three rows and three columns, with an overall regular square shape. Fig. 16(b) shows the imaging result of PFA-LOSPI, where the defocus and distortion of the placed target are so severe that they obscure the "river". Fig. 16(c) shows the imaging result of the proposed BS-PCS-PFA, it is shown that the targets locate in real position and are well focused. The imaging result of BPA is shown in Fig. 16(d).

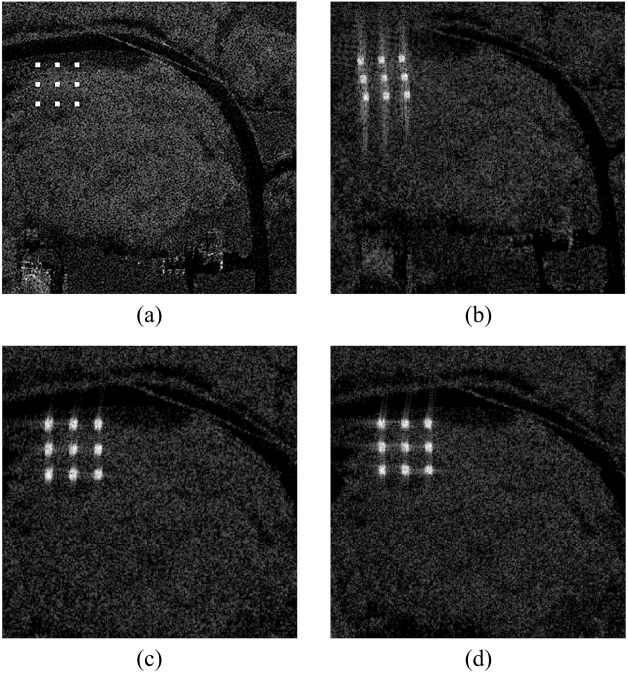

Fig.17. (a) Zoom-in image of input image. (b)-(d) Zoom-in imaging results of PFA-LOSPI, BS-PCA-PFA and BPA.

TABLE V
Quantitative Analysis of Extended Target Imaging Results

|  | Original Image | PFA-LOSPI | BS-PCS-PFA | BPA |
|---|---|---|---|---|
| Entropy | 4.20 | 6.21 | 5.16 | 5.58 |
| RMSE | 0 | 0.20 | 0.14 | 0.16 |
| PSNR | ∞ | 12.60 | 16.37 | 19.14 |
| SSIM | 1 | 0.17 | 0.22 | 0.27 |

The zoom-in imaging results are shown in Fig.17. Through the analysis of the strong point targets in the input image, it can be seen that the proposed method has a good effect on the compensation of WCE. The whole experiment analysis verifies the persistent imaging capability of the proposed method in different frequency bands. The quantitative results of the zoom-in imaging results are shown in TABLE V, including image entropy, root mean square error (RMSE), peak signal-to-noise ratio (PSNR), and structure similarity index measure (SSIM) [45], and the best value of each metric is the value of the original image. It can be seen that the proposed BS-PCS-PFA has the best performance in image entropy and RMSE, slightly inferior to BPA in PSNR and SSIM, but superior to PFA-LOSPI in all metrics.

*C. Large Observation Scenarios Experiments*

To verify the performance of the proposed method in different scenarios, a larger observation scenario, with a scene size of 500m*500m, is validated to demonstrate the real applicability of the proposed method. The system parameters are listed in Table VI. The imaging scene is shown in Fig. 18, and the interval of each point target is 100m. It can be calculated that the DiR is 27m, and the DeR is 382m for THz-band video SAR, while it is 80m for X-band video SAR. According to the sub-block imaging criterion of BS-PCS-PFA, the imaging scene is uniformly divided into 24*24 sub-blocks with an adjacent overlap of 6.43 m, and then the beam-segmenting fast filtering and sub-block imaging are performed for each sub-block data.

Moreover, motion errors have been added to validate the effectiveness of the three-step MoCo method. The platform vibration $\Delta R_e(t)$ is modeled as follows:

$$\Delta R_e(t) = \sum_{i=1}^{M} A_i \cdot \sin(2\pi f_i t + \varphi_i) \qquad (53)$$

TABLE VI
System Parameters for Large Observation Scenario Experiments

| Parameters | Values |
|---|---|
| Carrier frequency | 9.6GHz/220GHz |
| Reference slant range | 2500m |
| Bandwidth | 1.2GHz |
| Sampling frequency | 50MHz |
| Pulse repetition frequency | 25kHz |
| Sampling points | 4096 |

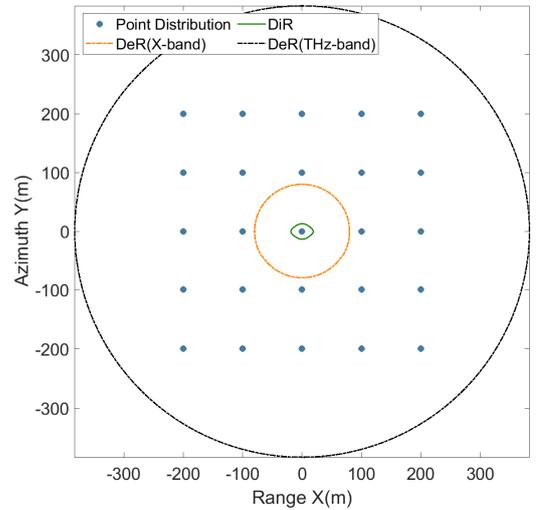

Fig. 18. Image scene of large observation scene experiments.

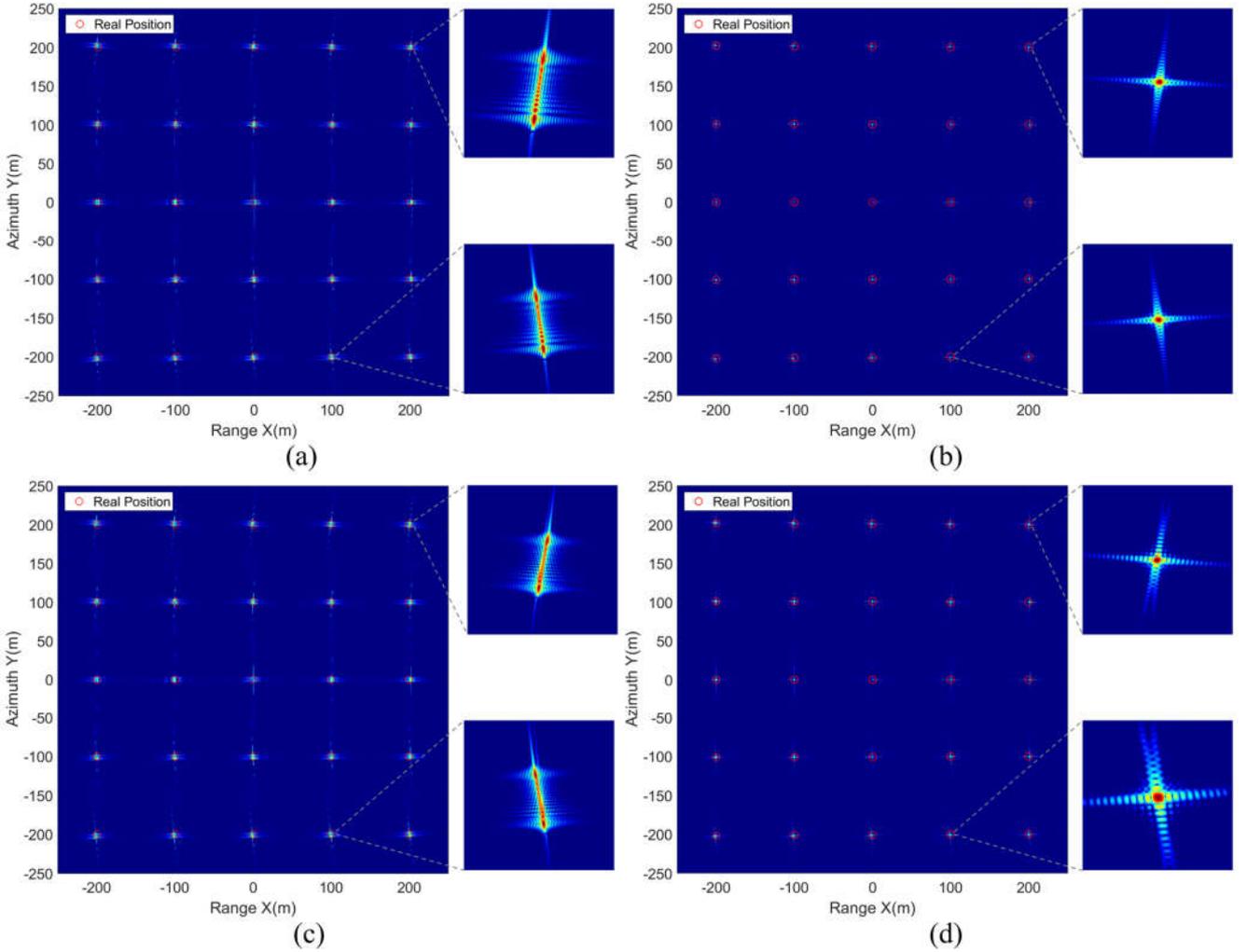

Fig.19. Imaging results of the proposed method in large observation scenarios with motion errors. (a)(b) Imaging results without MoCo and using three-step MoCo in THz-band. (c)(d) Imaging results without MoCo and using three-step MoCo in X-band.

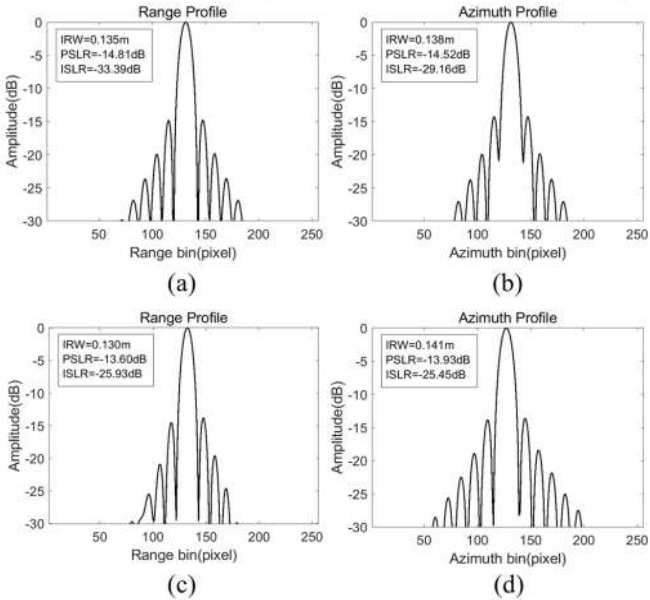

Fig.20. Image profiles of point target P1 processed by the proposed method. (a)(b) Range and azimuth profiles in THz-band. (c)(d) Range and azimuth profiles in X-band.

where M is the number of platform vibration components, $A_i$, $f_i$ and $\varphi_i$ represents the amplitude, frequency, and initial phase of the i-th vibration component, respectively. In this article, $M = 1$, the value of $A_i$ is 1.5 times the corresponding wavelength, $f_i = 5Hz$ and $\varphi_i = \pi/3$.

The final imaging results are shown in Fig. 19, where Fig. 19(a) and (b) show the imaging results of the proposed method in THz-band without using MoCo and using the three-step MoCo, respectively, and Fig. 19(c) and (d) show the corresponding imaging results of the proposed method in X-band, respectively. It can be seen that for the method that do not use MoCo, motion errors cause severe defocusing in its imaging results, while for method with three-step MoCo, the distortion or defocusing do not exist, and each target is accurately focused, which demonstrates the effectiveness of the three-step MoCo method.

The quantitative analysis of point target P1(200,200) is shown in Fig. 20, where Fig. 20(a) and (b) shows the range and azimuth profiles in THz-band, and Fig. 20(c) and (d) shows the range and azimuth profiles in X-band, respectively. From the image profiles and the corresponding point target quality metrics, it can be seen that the point target is well focused, which demonstrate the imaging performance of the proposed method in large scenarios.

## VI. CONCLUSION

In this paper, a BS-PCS-PFA is proposed for video SAR to improve the performance of traditional PFA. Firstly, to replace the wavenumber 2-D interpolation, the improved PCS for video SAR PFA are derived and the coarse image is obtained. Secondly, to eliminate the distortion and defocus in coarse image caused by the wavefront curvature error, the proposed beam-segmenting fast filtering and sub-block imaging method decomposes the coarse image into multiple sub-beam data and accomplishing down-sample at the same time, which can keep the complexity of algorithm not increasing too much. Then, the sub-beam data are processed by PFA, and then mosaicking the sub-images to obtain the full SAR image. Besides, since the imaging processes of sub-blocks do not affect each other, it is suitable for parallel processing in hardware to further accelerate the imaging process. The computational load of the BS-PCS-PFA and traditional methods are analyzed and verified by point target and extend target simulation experiment. It is shown that the BS-PCS-PFA behave better performance than traditional methods in terms of both processing efficiency and imaging quality. Moreover, the proposed method is a robust solution which can be easily extended to various working modes, such as linear spotlight mode and curve flight trajectories, as long as the platform position is accurately measured.

There are still some improvements for the proposed method, which are the focus of further research. For example, the proposed method is intended to be applicable to any frequency band, but in reality, THz video SAR is more sensitive to motion errors which may be introduced by airflow disturbance and attitude control. Therefore, the more accurate and efficient motion error compensation algorithm and its combination with the proposed method will be studied in the future. Then, they will be verified by the real data after the development of our video SAR system.